\PassOptionsToPackage{dvipsnames,table}{xcolor} 
\documentclass[letterpaper, 10 pt, conference]{ieeeconf}  

\IEEEoverridecommandlockouts                              

\makeatletter
\def\endthebibliography{%
	\def\@noitemerr{\@latex@warning{Empty `thebibliography' environment}}%
	\endlist
}
\makeatother

\usepackage{graphicx} 
\usepackage{mathptmx} 
\usepackage{times} 
\usepackage{amsmath} 
\usepackage{amssymb} 
\usepackage{siunitx} 
\usepackage{verbatim} 
\usepackage{mathrsfs} 
\usepackage{booktabs} 
\usepackage{multirow} 
\usepackage{upgreek} 
\usepackage{orcidlink} 

\title{\LARGE \bf  HIPPo: \underline{H}arnessing \underline{I}mage-to-3D \underline{P}riors for Model-free \\Zero-shot 6D \underline{Po}se Estimation}

\author{Yibo Liu$^{1,2,* \, \orcidlink{0000-0003-1143-3242}}$, \textit{Member, IEEE}, Zhaodong Jiang$^{1,3,*}$, Binbin Xu$^{1}$, Guile Wu$^{1}$, Yuan Ren$^{1 \, \orcidlink{0000-0002-4901-3596}}$, \\
Tongtong Cao$^{1}$, Bingbing Liu$^{1 \, \orcidlink{0000-0002-5272-3425}}$, Rui Heng Yang$^{1}$, Amir Rasouli$^{1}$, Jinjun Shan$^{2 \, \orcidlink{0000-0002-4911-6739}}$, \textit{Senior Member, IEEE}.
	\thanks{* Equal contribution. Work done during an internship at Noah's Ark Lab.}
    \thanks{$^{1}$ Huawei Noah Ark's Lab, Markham, Ontario  L3R 5A4, Canada.}
    \thanks{$^{2}$ York University, Toronto, Ontario M3J 1P3, Canada.}
    \thanks{$^{3}$ University of Toronto, Toronto, Ontario, M5S 1A1, Canada.} 
		\thanks{{\tt\scriptsize buaayorklau@gmail.com, zhaodong.jiang@mail.utoronto.ca, 
        \thanks{{\tt\scriptsize
        \{binbin.xu,rui.heng.yang2\}@h-partners.com}} }}
        \thanks{{\tt\scriptsize
        \{guile.wu, yuan.ren3, caotongtong, liu.bingbing,amir.rasouli\}@huawei.com,
        jjshan@yorku.ca
        } } }
\begin{document}

\maketitle
\thispagestyle{empty}
\pagestyle{empty}	
\begin{abstract}
This work focuses on model-free zero-shot 6D object pose estimation for robotics applications.
While existing methods can estimate the precise 6D pose of objects, they heavily rely on curated CAD models or reference images, the preparation of which is a time-consuming and labor-intensive process. Moreover, in real-world scenarios, 3D models or reference images may not be available in advance and instant robot reaction is desired.
In this work, we propose a novel framework named HIPPo, which eliminates the need for curated CAD models and reference images by harnessing image-to-3D priors from Diffusion Models, enabling model-free zero-shot 6D pose estimation.
Specifically, we construct HIPPo Dreamer, a rapid image-to-mesh model built on a multiview Diffusion Model and a 3D reconstruction foundation model.
Our HIPPo Dreamer can generate a 3D mesh of any unseen objects from a single glance in just a few seconds.
Then, as more observations are acquired, we propose to continuously refine the diffusion prior mesh model by joint optimization of object geometry and appearance.
This is achieved by a measurement-guided scheme that gradually replaces the plausible diffusion priors with more reliable online observations.
Consequently, HIPPo can instantly estimate and track the 6D pose of a novel object and maintain a complete mesh for immediate robotic applications.
Thorough experiments on various benchmarks show that HIPPo outperforms state-of-the-art methods in 6D object pose estimation when prior reference images are limited. 
The project page is: https://hippope.github.io/

%
\end{abstract}

\section{INTRODUCTION} \label{intro}

\begin{figure}[th]
	\centering
	\includegraphics[width=3.3in]{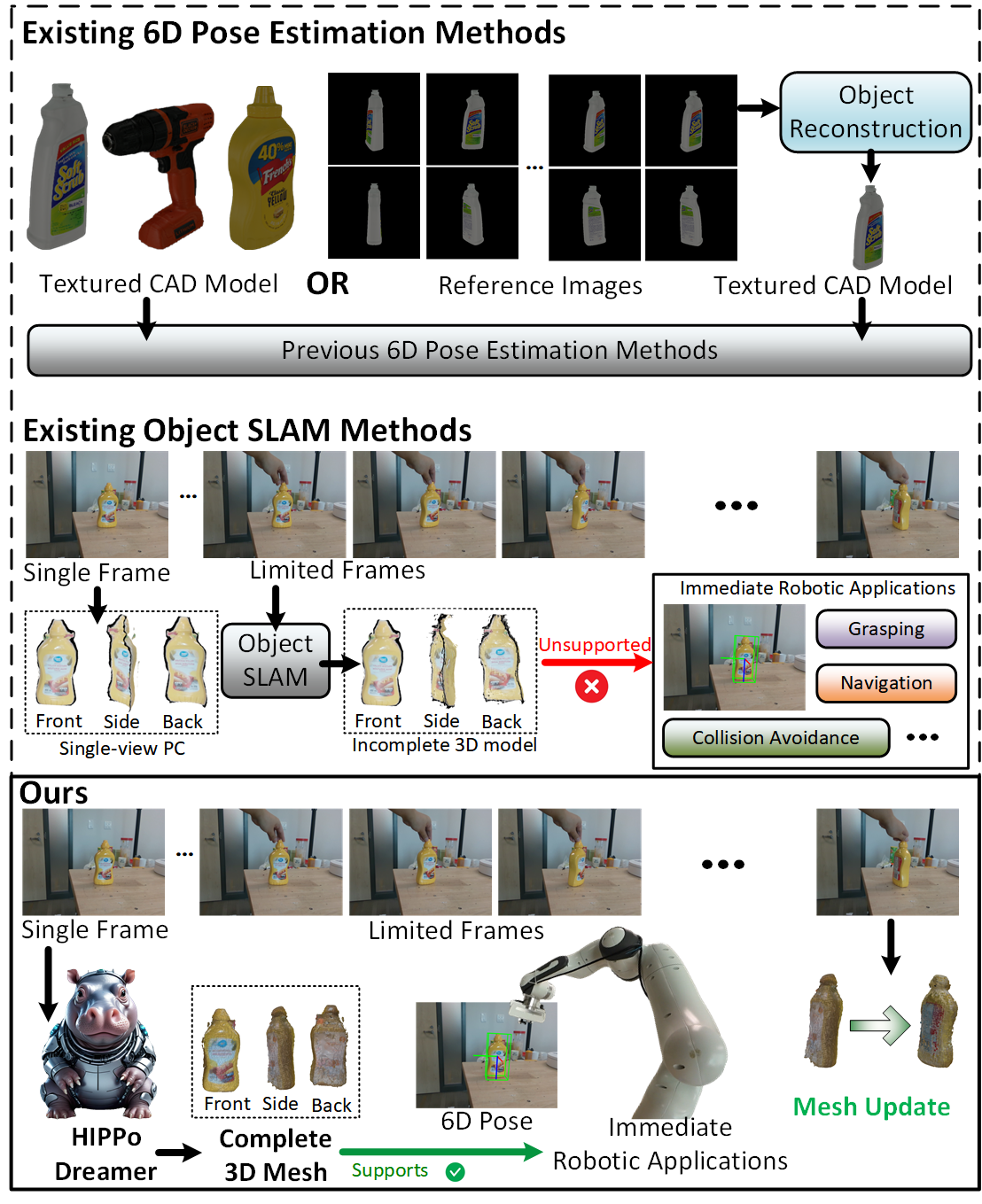}
	\caption{
   Compared to existing SOTA 6D pose estimation methods \cite{fp,sam6d,gigapose}, HIPPo eliminates the need for a textured 3D model or reference images in advance, while also optimizing the reference 3D model online. Compared to existing object SLAM methods \cite{xu2019mid, bundlesdf}, HIPPo sustains a complete 3D model from the first glance of the object, enabling immediate robotic applications.
    }
	\label{fig1}
    \vspace{-0.3in}
\end{figure}
6D pose estimation \cite{fp,gigapose,sam6d} is crucial for robotic applications \cite{navigation,liao,shuo2} such as grasping, navigation, exploration, and collision avoidance.
Although many 6D pose estimation methods \cite{fp,gigapose,sam6d,onepose,onepose++,fs6d} exist, as shown in Fig. \ref{fig1}, they often demand the textured CAD model of the object in advance.
Crafting a curated CAD model is time-consuming and labor-intensive. Thus, some research \cite{fp,onepose++,fs6d,onepose} focuses on using reference images or a video of the object as input instead of a 3D model.
Nevertheless, many of them \cite{fp,onepose,onepose++} still need to perform object reconstruction \cite{bundlesdf,colmap} to transform the reference images into a textured 3D model or require posed images of the object as a reference \cite{sam6d,fs6d}.   
Unfortunately, in real-world robotic applications, 3D models or reference images may not be available a priori, limiting the system’s deployment in open-world scenarios where object models are not always accessible. Recently, image-to-3D methods \cite{wonder3d,zero123++,instantmesh,vqadiff} have shown robust zero-shot prediction capabilities. Specifically, Diffusion Models \cite{zero123++,wonder3d} trained on the large-scale dataset \cite{obj} can render novel views of arbitrary unseen objects. Inspired by this, we aim to harness the learned image-to-3D priors from Diffusion Models \cite{zero123++,wonder3d} to boost 6D pose estimation without relying on CAD models or reference images. \par
Yet this task is challenging due to the two limitations of existing image-to-3D methods \cite{wonder3d,zero123++,instantmesh,vqadiff}. First, 6D pose estimation \cite{fp} requires the reference 3D model to have the same scale as the real-world object, whereas image-to-3D methods do not account for the scale of the generated model. Second, the image-to-3D problem is inherently ill-posed \cite{vqadiff}: given a conditioned image, the uncaptured views can exhibit many plausible appearances and geometries, causing discrepancies between the generated 3D model and the actual views. Unfortunately, existing image-to-3D methods focus solely on generating a complete 3D model, rather than adapting to new observations.
In addition, 6D pose estimation methods \cite{fp, onepose++, onepose} treat object reconstruction as a preprocessing step for pose estimation and do not consider the optimization of the reference model during pose estimation. Although the object SLAM methods \cite{xu2019mid, bundlesdf} can conduct simultaneous 6D pose estimation and model refinement, as shown in Fig. \ref{fig1}, they often produce incomplete models from limited views, making their models less suitable for immediate robotic applications. 
\par
In this work, we propose HIPPo, a novel framework that leverages image-to-3D priors for model-free zero-shot 6D pose estimation. 
As shown in Fig. \ref{fig1}, compared to existing 6D pose estimation methods \cite{fp,gigapose,sam6d}, HIPPo eliminates the need for preparing a textured CAD model in advance.
It can be initialized from any first glance of the object and simultaneously estimates the 6D pose while optimizing the 3D model of the object online.
For the initialization of HIPPo, we design HIPPo Dreamer, a rapid image-to-mesh strategy based on a multiview Diffusion Model \cite{wonder3d} and a 3D reconstruction foundation model \cite{mast3r}. It generates a 3D mesh of an unseen object from a single reference image in just a few seconds and can recover the mesh's physical scale.
Recognizing that the diffusion-based mesh has limited reliability in uncaptured views due to the ill-posed nature \cite{make}, we further develop a measurement-guided algorithm to continuously optimize the mesh. 
Specifically, a viewpoint sphere tracks the relative pose changes between frames. Mesh optimization is triggered when a keyframe is recognized to replace the diffusion prior with more reliable online measurements of appearance and geometry. 
We conduct extensive experiments on various challenging benchmarks.
Experimental results demonstrate that the proposed method significantly outperforms state-of-the-art (SOTA) 6D pose estimation methods \cite{fp,gigapose,sam6d} when prior reference images are limited.
\par
The \textbf{contributions} of this work are:
\begin{itemize}
\item We develop a novel model-free zero-shot 6D pose estimation framework named HIPPo. Compared to existing 6D pose estimation methods \cite{fp, gigapose, sam6d}, it eliminates the need for a textured CAD model or for dense posed reference images.
\item We propose an instant image-to-mesh strategy called HIPPo Dreamer. Compared to InstantMesh \cite{instantmesh}, a popular instant image-to-mesh solution, HIPPo Dreamer introduces the new capability of scale recovery. Compared to BundleSDF, the SOTA object SLAM method, HIPPo Dreamer always maintains a complete 3D mesh from the first glance.
\item We design a measurement-guided method to optimize the mesh online, enhancing mesh fidelity with more observations and surpassing the diffusion prior mesh.
\end{itemize}

\section{Related Work}
\label{rel}
\noindent\textbf{Instance-level Image-to-3D Methods.}
Instance-level image-to-3D approaches \cite{wonder3d, zero123++,instantmesh,vqadiff} aim to generate 3D representations from a single image. Specifically, Diffusion Models \cite{zero123++, wonder3d} have demonstrated strong zero-shot prediction abilities, benefiting from training on large-scale datasets such as Objaverse \cite{obj}. Since reconstruction quality is usually prioritized over efficiency, they take several minutes and even tens of minutes to reconstruct 3D models, which limits their applicability in real-time scenarios. While some methods \cite{instantmesh,vqadiff} can achieve faster image-to-3D generation, they generally do not incorporate incremental optimization of the 3D model.\\
\noindent\textbf{6D Pose Estimation Methods.}
The implementation \footnote{Their pose estimation networks require dense posed reference images, which are sampled from a textured CAD model in their pipelines.} of most existing 6D pose estimation methods, such as FoundationPose \cite{fp}, GigaPose \cite{gigapose}, and SAM6D \cite{sam6d}, requires a textured CAD model in advance, which requires intensive time and labor to craft \cite{lpr}. For example, when handling an unseen object, FoundationPose \cite{fp} requires running BundleSDF \cite{bundlesdf} to generate the reference 3D model, which takes tens of minutes to complete. Similarly, OnePose \cite{onepose} and OnePose++ \cite{onepose++} propose recording a video scan of the object and utilizing Structure-from-Motion \cite{colmap} to reconstruct the object. FS6D \cite{fs6d} does not reconstruct the object from reference images but requires reference images with pose labels. However, the prior operation involving the reference 3D model or reference image sampling may be impractical in real-world settings, especially when instant robotic action is required at first sight of the object. Recently, Zero123-6D \cite{zero1236d} proposes leveraging the Diffusion Model for 6D pose estimation. However, the category-level pose estimation strategy of Zero123-6D does not fully utilize the instance-level object generation ability of Diffusion Models \cite{wonder3d,zero123++} and, like previous methods \cite{fp,gigapose,sam6d,onepose,onepose++,fs6d}, does not consider further model optimization during pose estimation. In contrast, object-SLAM methods \cite{xu2019mid, bundlesdf} reconstruct objects in real-time without prior knowledge, tracking and optimizing the geometry \cite{mvdeepsdf} and appearance \cite{yanglearning,unigaussian} from scratch. However, they struggle to provide complete models when observations are scarce. Incorporating advancements in object generation, the object-level mapping proposed in \cite{xu2022learning} introduces DeepSDF-like~\cite{mvdeepsdf} generative priors to constrain object shapes, enabling the estimation of complete shapes and poses under occlusions. However, it is limited to a single category due to the shortcomings of these generative priors. Concurrently, GOE \cite{liao2024toward} extends this approach by leveraging a multi-category 3D diffusion prior, but its optimization efficiency remains limited.

\begin{figure*}
	\centering
	\includegraphics[width=17.5cm]{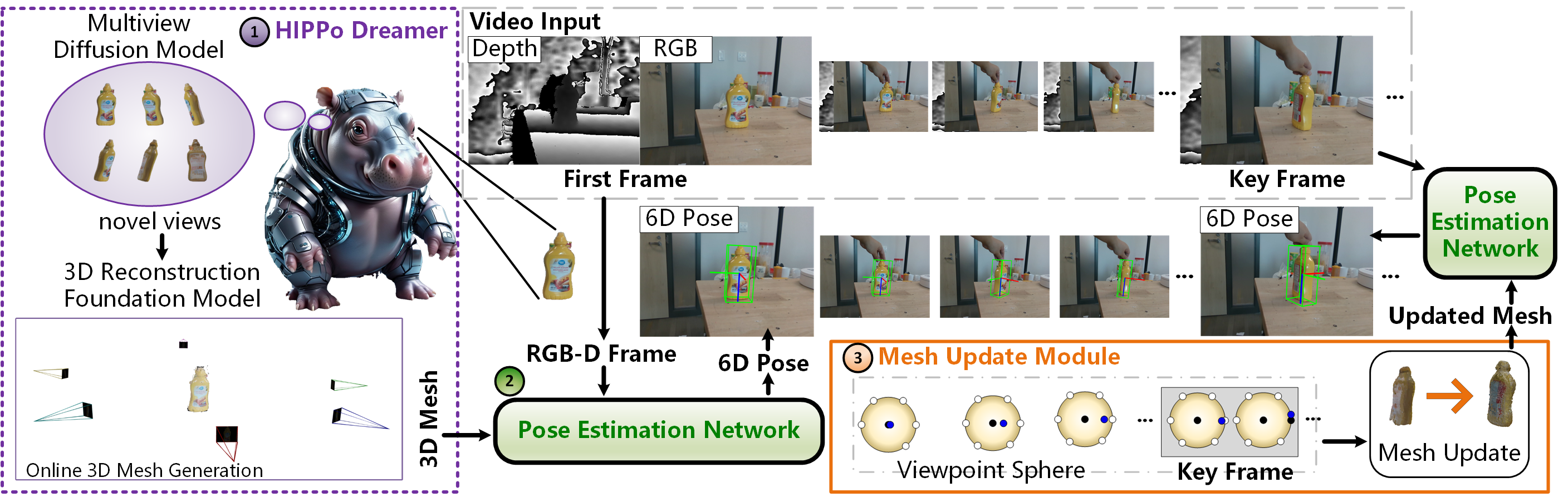}
	\caption{\textbf{Overview of HIPPo.} Given a video consisting of RGB-D frames, Grounding DINO \cite{grounding} is first applied to segment the object based on a prompt. 
    Next, the proposed HIPPo Dreamer, built on a multiview Diffusion Model and a 3D reconstruction foundation model, generates a 3D mesh of the object from the first detected frame in a few seconds. 
   Then, the diffusion prior mesh is provided to the pose estimation network to estimate the 6D pose in real time.
   Meanwhile, the mesh optimization module monitors viewpoint changes through a predefined viewpoint sphere and triggers mesh optimization when the viewpoint varies dramatically. The module then replaces the diffusion prior with more reliable appearance and geometry from online measurements.
     }
	\label{overview}
  
\end{figure*}
\section{Methodology}
An overview of the proposed HIPPo is shown in Fig. \ref{overview}. It consists of three components: the HIPPo Dreamer, the pose estimation network, and the mesh optimization module. The HIPPo Dreamer, described in Sec. \ref{secdreamer}, is responsible for initializing HIPPo by instantly generating a 3D mesh from the detected first frame. The 6D pose estimation network, introduced in Sec. \ref{6dpose}, estimates the 6D pose based on the reference mesh. The mesh update module, detailed in Sec. \ref{meshupdate}, refines the mesh by updating the diffusion prior with more reliable online measurements.
\subsection{HIPPo Dreamer} \label{secdreamer}
\subsubsection{Scale Recovery Problem Analysis}
As introduced in Sec. \ref{intro}, the correct scale of the reference 3D model is the prerequisite for 6D pose estimation. In this work, scale recovery refers to finding the constant scale, $s$, represented by:
\begin{equation}
s = g_{max}/r_{max} \label{scale}
\end{equation}
where $g_{max}$ denotes the maximum side length of the oriented bounding box (OBB) \cite{obb} of the generated model, while $r_{max}$ represents that of the real-world object. Suppose that the camera intrinsics are known, $r_{max}$ can be obtained at the first frame through depth measurement after object segmentation. But it should be noted that $r_{max}$ corresponds to the OBB that encloses the partial point cloud of the real-world object captured from the view of the first frame, denoted by the original view. Therefore, to recover the scale, we need to compute $g_{max}$ by leveraging the estimated depth of the original view. For this reason, InstantMesh \cite{instantmesh}, though addressing the instant image-to-mesh problem, is not applicable here as it generates fixed views independent of the original view and does not provide direct depth estimation for it. Consequently, it is necessary to develop a new image-to-mesh framework that supports scale recovery. \\
%
\subsubsection{Framework Design} \label{frameworkdesign} 
\begin{figure}[th]
	\centering
	\includegraphics[width=2.0in]{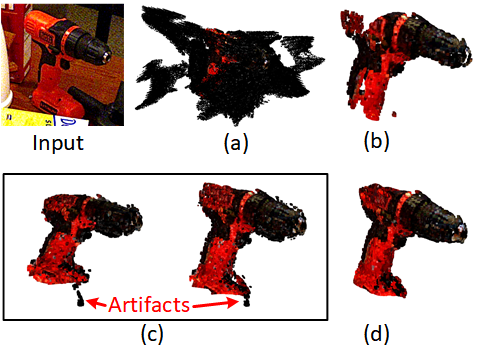}
	\caption{ Comparison of the vanilla MASt3R (a)(b)(c) and our modified MASt3R (d). (a): A low threshold preserves too many background points. (b): A high threshold results in an incomplete model by masking out some foreground object points. (c): Even with careful fine-tuning, artifacts may remain around the object, affecting the judgment of its scale. (d): Our modified MASt3R generates artifact-free 3D models without requiring fine-tuning of the hyperparameter.
    }
	\label{mast3rnew}
\end{figure}

\textbf{(1) Object Segmentation.} Given a frame of an unseen object, we use a guiding prompt and Grounding DINO \cite{grounding} to segment the object first from the RGB image and then use the same mask to segment the object from the depth image. \textbf{(2) Instant Image-to-multiview-to-mesh Generation.} Inspired by InstantMesh \cite{instantmesh}, we apply an image-to-multiview-to-mesh strategy. In particular, we first employ the multiview Diffusion Model \footnote{Although Wonder3D \cite{wonder3d} provides a complete solution for image-to-mesh, we only adopt its Diffusion Model for two reasons. First, its multiview-to-mesh step takes several minutes, which is not instant. Second, it does not directly provide depth estimation of the first (original) view.} from Wonder3D \cite{wonder3d} for image-to-multiview generation. This is because the first view among the multiple views generated by Wonder3D always matches the original view. If the depth of the first view is estimated, scale recovery follows from Eq. (\ref{scale}). Thereafter, we adopt a modified MASt3R \cite{mast3r}, a 3D reconstruction foundation model, for simultaneous depth estimation for scale recovery and multiview-to-mesh conversion. We modify MASt3R since it considers the background in optimization and relies on a hyperparameter, \textit{the minimum confidence threshold}, to remove the background from the 3D reconstruction result. 
Unfortunately, as shown in Figs. \ref{mast3rnew}(a)(b)(c), this hyperparameter requires careful fine-tuning for each object.
To address this problem, we modify the background point masking process in the vanilla MASt3R by integrating SAM \cite{sam} into the process to generate accurate object masks, instead of simply masking out all points with matching confidence below the threshold. Then, to further remove the artifacts, we apply the Statistical Outlier Removal (SOR) filter, defined as  
\begin{equation}
\|\mathbf{p} - \mu\| > k \sigma \label{sor}
\end{equation}
where \(\mathbf{p}\) is a point in the cloud, \(\mu\) is the mean position of its \(N\) nearest neighbors, \(\sigma\) is the standard deviation of distances, and \(k\) denotes threshold (\(k=300\) in practice).  Accordingly, as shown in Fig. \ref{mast3rnew}(d), our modified MASt3R can robustly generate 3D point clouds free of background points, without requiring hyperparameter fine-tuning.
\noindent{\textbf{(3) Scale Recovery.}} An estimated partial point cloud of the object can be obtained through the estimation of depth and intrinsics of the original view by MASt3R. By computing the OBB of it, $g_{max}$, as shown in Eq. (\ref{scale}), is acquired and the scale is determined through through Eq. (\ref{scale}). 
However, we experimentally found that for both measured and estimated depth images, even when employing SAM \cite{sam} to provide object masks, the partial point clouds of the object remain noisy, particularly around the edges, affecting the values of $r_{max}$ and $g_{max}$. 
Therefore, we apply the SOR filter, as shown in Eq. (\ref{sor}), to denoise both the measured and estimated point clouds before computing the scale.
Although straightforward, this SOR denoising step is effective and necessary for scale recovery. Finally, we convert the scaled 3D point cloud into a 3D mesh using Poisson surface reconstruction \cite{poisson}.
\par
\begin{figure}[t]
	\centering
	\includegraphics[width=3.3in]{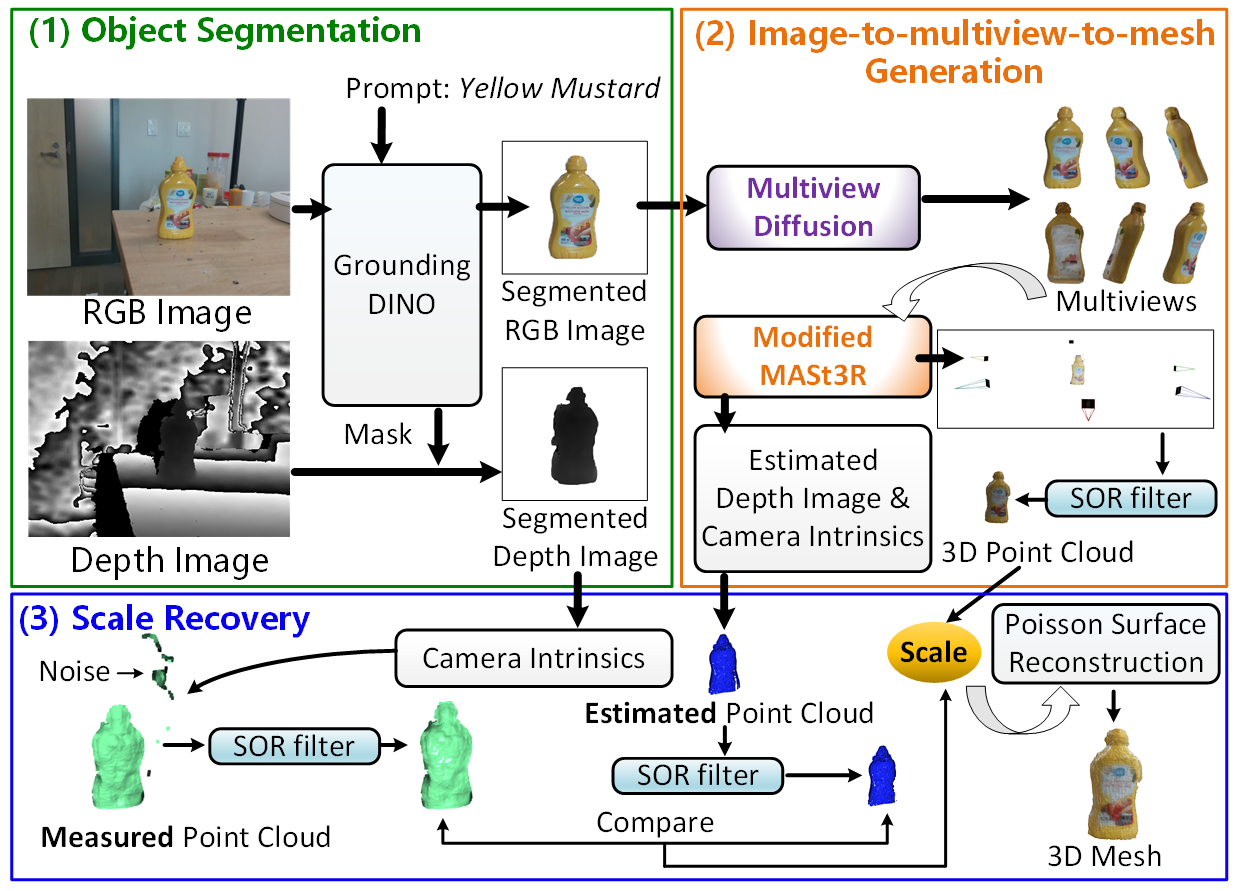}
	\caption{An illustration of the HIPPo Dreamer pipeline.
    }
	\label{dreamer}
    
\end{figure}

In summary, Fig. \ref{dreamer} shows the major steps of the proposed HIPPo Dreamer.

\subsection{6D Pose Estimation} \label{6dpose}
Once a textured 3D mesh is generated by the proposed HIPPo Dreamer, we employ the Pose Refinement network from FoundationPose \cite{fp} for 6D pose estimation due to its SOTA performance. The network takes two inputs: one is a rendering of the generated object conditioned on the most recent pose estimate, and the other is a cropped observation from the camera.
The Siamese Network comprises two feature embedding networks with shared weights that extract feature maps from the two RGB-D input branches.
These feature maps are then concatenated and fed into additional CNN blocks, where they are tokenized by dividing them into patches with positional embeddings.
A transformer then predicts a pose update, iteratively refining the pose estimation over a few iterations.

\subsection{Mesh Update Module} \label{secmeshupdate}
Due to the ill-posed nature of the image-to-3D problem, HIPPo Dreamer may generate varying predictions of unseen views with low-fidelity rendering and geometry.
%
This is unfavorable for 6D pose estimation, as the inconsistency between the reference model and the real-world object disrupts the matching between the rendered and measured views (see Sec. \ref{6dpose}). Thus, as more observations become available, it is desirable to update the reference mesh. To enable rapid reference mesh updates, a problem not considered in previous pose estimation and image-to-3D research \cite{fp, gigapose, sam6d, wonder3d, instantmesh, vqadiff}, we propose using viewpoint variation as a trigger for mesh updates. 
\par

\begin{figure}[t]
	\centering
	\includegraphics[width=3.3in]{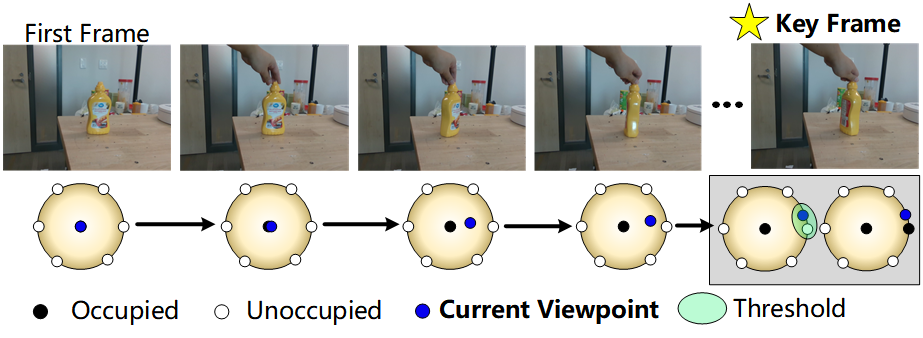}
	\caption{An illustration of the viewpoint sphere. Each circle on the sphere represents a viewpoint. By monitoring the current viewpoint on the sphere, key frames representing dramatic viewpoint changes are recognized, triggering a mesh update at these key frames.}
	\label{sphere}
\end{figure}
\begin{figure}[t]
	\centering
	\includegraphics[width=3.3in]{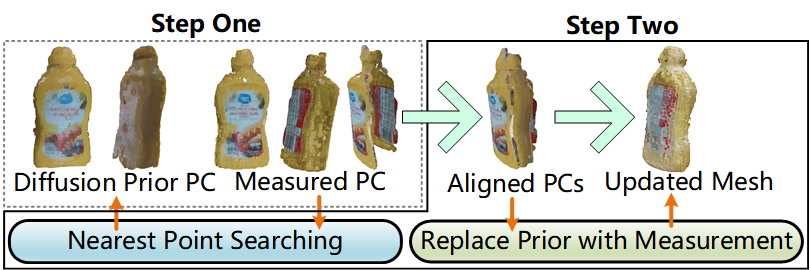}
	\caption{An illustration of the proposed mesh update method. It consists of two steps. In step one, the measured point cloud and diffusion prior point cloud are aligned. In step two, the diffusion prior is replaced with measurements based on nearest point searching.}
	\label{meshupdate}
   
\end{figure}

We first design a viewpoint sphere to select new perspectives with significant viewpoint shifts and filter out redundant poses.
As shown in Fig. \ref{sphere}, there are $N$ viewpoints uniformly distributed on the viewpoint sphere. Then, frames are classified as keyframes for updates when they align with unoccupied viewpoints. We set a tolerance threshold to account for slight misalignments, ensuring effective yet controlled updates.
The pose of the first frame is aligned with an arbitrary viewpoint on the sphere and we only monitor the relative rotation between frames. 
The corresponding viewpoint on the sphere is marked as occupied after conducting mesh update and will not trigger further mesh updates. \par
Next, we propose a mesh update method based on the modality of the 3D colored point cloud, as shown in Fig. \ref{meshupdate}. \textbf{(1) Step One.} We register all the measured points into the first frame and apply the SOR filter to denoise the point cloud. Then, we transform the measured points into the object frame using the pose estimation result of the first frame. In this way, as seen in step one of Fig. \ref{meshupdate}, both the 3D models provided by HIPPo Dreamer and the online measurements are registered in the object frame, with their scales matched in Sec. \ref{frameworkdesign}.  
\textbf{(2) Step Two.} We further build a KDTree on the diffusion prior 3D point cloud. 
For each point in the measured point cloud, we search for the nearest point in the diffusion prior mesh and replace the 3D position and color of the prior point with those of the measured point.
Thereafter, the updated 3D colored point cloud is transformed into a mesh using Poisson surface reconstruction \cite{poisson}. In practice, we apply Farthest Point Sampling \cite{mvdeepsdf} to downsample the measured point cloud if the number exceeds 30,000 to ensure that the mesh update can be completed within a few seconds.
\begin{figure}[t]
	\centering
	\includegraphics[width=3.0in]{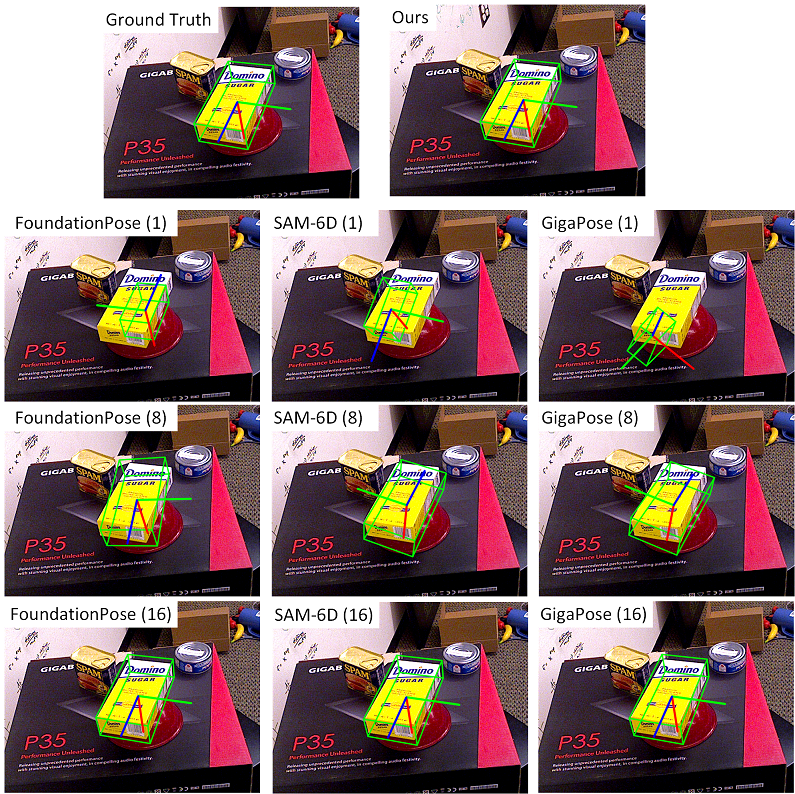}
	\caption{Qualitative comparison of the proposed method against SOTA methods on the YCB-Video dataset \cite{ycbv}. The number in brackets refers to the number of reference images known in advance.}
	\label{comsota}
   
\end{figure}
\section{Experiments} \label{exp}
\subsection{Comparison with SOTA 6D Pose Estimation Methods} \label{com1}
\noindent{\textbf{Benchmarks and Experimental Setup.}} In this section, we the 6D pose estimation performance of HIPPo on two popular benchmarks: YCB-Video \cite{ycbv} and LM-O \cite{lmo}. In particular, for each object, we have three levels of reference images: 1, 8, and 16. One reference image represents a single glance of the object. For a fair comparison, the reference image provided to all methods is taken from the object in the first detected frame. Next, using the ground truth object pose from the first frame and the ground truth object model, we render 8 and 16 images by rotating the virtual RGB-D camera around the z-axis of the object frame in Blender \cite{blender}. Then, we apply BundleSDF \cite{bundlesdf} to reconstruct the textured CAD models from the reference images. Following \cite{fp}, 16 reference images are sufficient to construct a complete model, while the 3D shape is relatively incomplete when reconstructed from 8 images and extremely incomplete when using just 1 image. We set 36 key points on the viewpoint sphere (see Fig. \ref{sphere}). \par
\noindent{\textbf{Competitors and Metrics.}} We compare HIPPo with three SOTA 6D pose estimation methods: FoundationPose \cite{fp}, SAM-6D \cite{sam6d}, and GigaPose \cite{gigapose}. The implementations of the competitors require textured CAD models. Following \cite{fp}, we use the Area Under the Curve (AUC) of ADD and ADD-S \cite{ycbv} as metrics to evaluate 6D pose estimation performance.
\\
\noindent{\textbf{Experimental Results and Analysis.}}
Qualitative and quantitative comparisons of our approach against SOTA methods \cite{fp,sam6d,gigapose} on YCB-Video \cite{ycbv} are presented in Fig. \ref{comsota} and Table \ref{ycbtab}, respectively.
As shown, although the SOTA methods \cite{fp,sam6d,gigapose} yield very promising results when the number of reference images is 16, indicating that a complete reference mesh is available, their performance dramatically degrades as the number of reference images decreases.
In contrast, our method, which always uses only one reference image, is slightly inferior to FoundationPose \cite{fp} and SAM-6D \cite{sam6d}, but outperforms GigaPose \cite{gigapose} when they are provided with 16 reference images.
However, when reference images are insufficient, which is expected to occur in immediate robotic applications, the proposed method demonstrates significantly superior performance over the SOTA competitors.

\begin{table}[t]
\caption{Quantitative comparison with SOTA methods on YCB-Video \cite{ycbv}. Img. refers to the number of reference images. }
\centering
	\resizebox{1.0\columnwidth}{!}{
\begin{tabular}{c|c|c|c|c|c|c|c|c|c|c}
\hline
\multirow{2}{*}{Method} &  \multicolumn{3}{c|}{GigaPose}    & \multicolumn{3}{c|}{SAM-6D}  & \multicolumn{3}{c|}{FoundationPose}  &  \multirow{2}{*}{\textbf{Ours}}   \\ 
 &  \multicolumn{3}{c|}{\cite{gigapose}}    & \multicolumn{3}{c|}{\cite{sam6d}}  & \multicolumn{3}{c|}{\cite{fp}}  &     \\ \hline
Img. & \cellcolor{green!30}1 &8 & 16 & \cellcolor{green!30}1 &8 & 16 & \cellcolor{green!30}1 &8 & 16 &\cellcolor{green!30}1 \\ \hline
ADD  & \cellcolor{green!30}16.42 & 24.70 & 61.71 & \cellcolor{green!30}23.85 &38.67 & 90.06 & \cellcolor{green!30}35.32 &51.59 & 96.56 &\cellcolor{green!30}\textbf{89.07} \\ \hline
ADD-S & \cellcolor{green!30}32.85 &43.73 & 82.23 & \cellcolor{green!30}40.62 &71.28 & 98.54 & \cellcolor{green!30}79.71 &90.24 & 99.47 &\cellcolor{green!30}\textbf{97.00} \\ \hline
\end{tabular}
}
\label{ycbtab}
\end{table}

\begin{table}[t]
\caption{Quantitative comparison with SOTA methods on LM-O \cite{lmo}. Img. refers to the number of reference images.}
\centering
	\resizebox{1.0\columnwidth}{!}{
\begin{tabular}{c|c|c|c|c|c|c|c|c|c|c}
\hline
\multirow{2}{*}{Method} &  \multicolumn{3}{c|}{GigaPose}    & \multicolumn{3}{c|}{SAM-6D}  & \multicolumn{3}{c|}{FoundationPose}  &  \multirow{2}{*}{\textbf{Ours}}   \\ 
 &  \multicolumn{3}{c|}{\cite{gigapose}}    & \multicolumn{3}{c|}{\cite{sam6d}}  & \multicolumn{3}{c|}{\cite{fp}}  &     \\ \hline
Img. & \cellcolor{green!30}1 &8 & 16 & \cellcolor{green!30}1 &8 & 16 & \cellcolor{green!30}1 &8 & 16 &\cellcolor{green!30}1 \\ \hline
ADD  & \cellcolor{green!30}15.87 & 30.62 & 66.53 & \cellcolor{green!30}21.16 &37.00 & 87.53 & \cellcolor{green!30}34.64 &42.64 & 92.20 &\cellcolor{green!30}\textbf{88.49} \\ \hline
ADD-S & \cellcolor{green!30}31.89 &54.45 & 87.45 & \cellcolor{green!30}39.70 &70.35 & 93.00 & \cellcolor{green!30}68.16 &82.23 & 97.17 &\cellcolor{green!30}\textbf{92.92} \\ \hline
\end{tabular}
}
\label{lmotab}

\end{table}

\par
The quantitative comparison of our approach against the SOTA methods \cite{fp,sam6d,gigapose} on LM-O \cite{lmo} is presented in Table \ref{lmotab}. From these results, we can also observe that these model- or reference-image-based methods heavily rely on the complete reference model, without which the performance degrades notably. In comparison, our approach remarkably outperforms the SOTA competitors when reference images are scarce, showing promising potential to facilitate instant applications. Despite the decent results, we also found that severe object occlusion can hinder the performance of our approach. A discussion of is provided in Sec. \ref{seclim}.

\subsection{Comparison with BundleSDF} \label{objre}
\begin{figure}[t]
	\centering
	\includegraphics[width=3.0in]{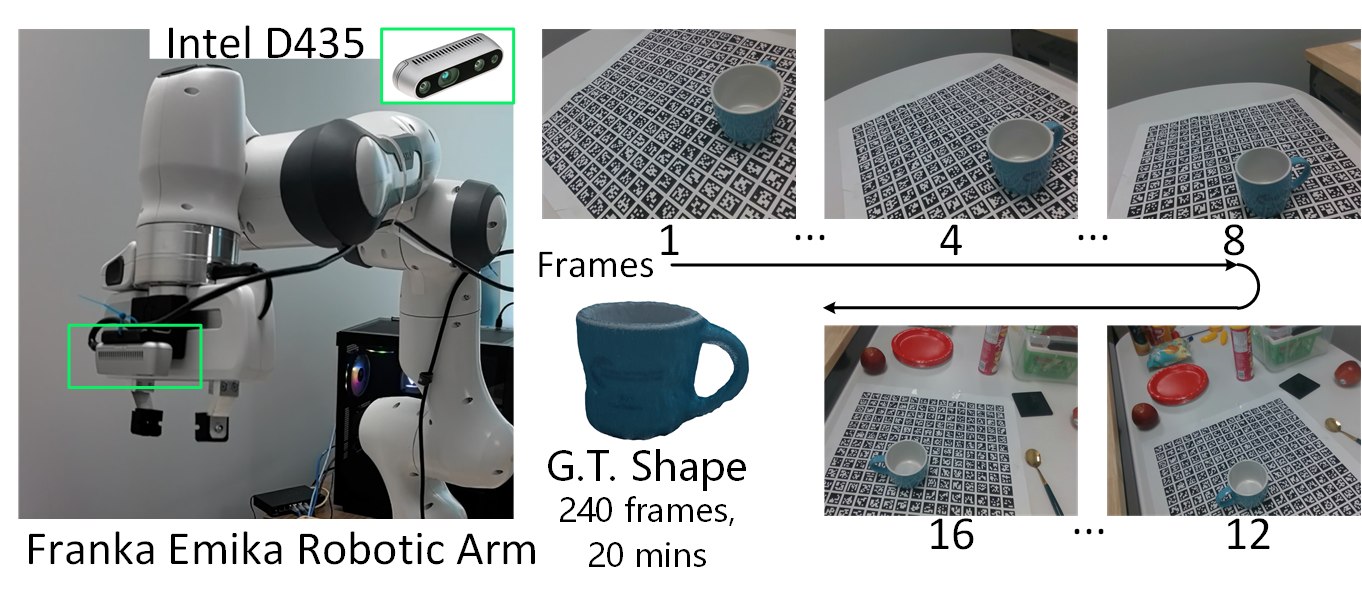}
	\caption{An illustration of the experimental setup. The frames are captured by an Intel D435 RGB-D camera (1280 $\times$ 720) mounted on a 7-DoF Franka Emika robotic arm.}
	\label{setup}
    
\end{figure}

\noindent{\textbf{Benchmark and Experimental Setup.}} In this section, we test HIPPo on a custom dataset to evaluate its object reconstruction quality and efficiency. In particular, this dataset contains RGB-D frames of four objects: three rendered using Blender \cite{blender} and one collected in the real world. The first three objects, $\mathrm{YCB}_{3}$ (sugar box), $\mathrm{YCB}_{5}$ (mustard bottle), and $\mathrm{YCB}_{15}$ (power drill), belong to the YCB object set \cite{ycb}. 
They are rendered with a virtual RGB-D camera at a resolution of 512 $\times$ 512, following a loop trajectory to capture 16 evenly spaced frames per object. The ground truth shapes are their CAD models from the YCB object set \cite{ycb}.
The experimental setup to scan the real-world object is shown in Fig. \ref{setup}. The ground truth shape of the mug is constructed using BundleSDF \cite{bundlesdf} with dense observations consisting of 240 RGB-D frames, taking approximately 20 minutes to complete. In particular, the robotic arm follows multiple pre-designed ring-view trajectories to scan the mug at different relative altitudes, completing the first 360-degree loop at the 16th frame.
Moreover, the rendered depth images are noise-free, while the real-world sampled depth images are noisy. The experiments are conducted using an NVIDIA GeForce RTX 3090 24GB GPU. \par
\begin{figure}[t]
	\centering
	\includegraphics[width=2.8in]{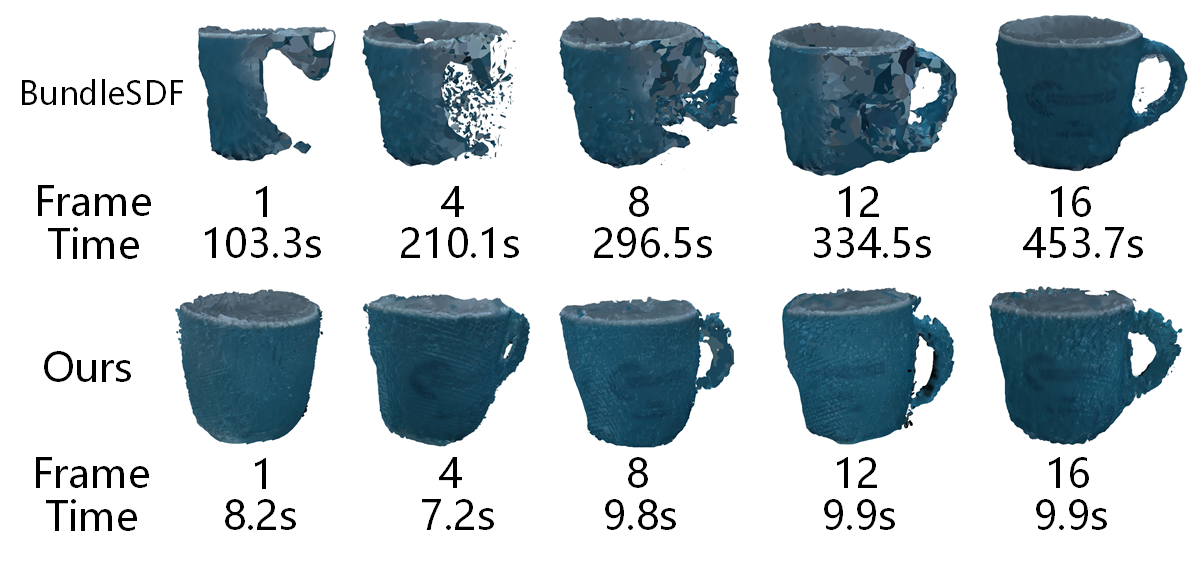}
	\caption{Qualitative comparison with BundleSDF \cite{bundlesdf} under different numbers of frames.}
	\label{sdf}
  
\end{figure} \par
\noindent{\textbf{Competitor and Metrics.}} We compare HIPPo against the SOTA object SLAM method, BundleSDF \cite{bundlesdf}. Moreover, for rendered RGB-D frames, the ground truth object poses are perfectly known and provided to BundleSDF. For real-world sampled RGB-D frames, the relative poses used by BundleSDF are obtained from visual odometry based on fiducial markers \cite{intensity,shuo1,liao,ghost,shuo3}. In contrast, HIPPo always uses its own pipeline to estimate relative poses. We apply the Chamfer Distance (CD) to evaluate the quality of the reconstructed object. The objects are normalized into a unit sphere before computing the CD, and the CD values are multiplied by $10^{3}$ for display.\\
\noindent{\textbf{Experimental Results and Analysis.}}
The qualitative and quantitative comparisons of HIPPo against BundleSDF \cite{bundlesdf} are presented in Fig. \ref{sdf} and Table \ref{sdftab}, respectively. Specifically, Fig. \ref{sdf} shows that our method always maintains a complete 3D model from the first frame compared to BundleSDF, with significantly better efficiency. Table \ref{sdftab} demonstrates that the 3D model fidelity of our method is superior to that of BundleSDF when the number \footnote{Note that the number of frames is a different concept from the number of reference images in Sec. \ref{com1}. Frames are the consecutive RGB-D frames provided to the methods. Reference images are known images used to reconstruct the model prior to 6D pose estimation.} of frames is fewer than 16. 
The inference time of our method on the first frame is always 8s, as the proposed HIPPo Dreamer processes a normalized input through a fixed pipeline. Our inference time on other frames is fewer than 10s since we downsample the accumulated measured point cloud to 30,000 points if it exceeds this number. As introduced in Sec. \ref{secmeshupdate}, the only time-consuming process in mesh update is the KDTree-based nearest point searching. An ablation study regarding this is provided in Sec. \ref{ab}.
Moreover, the continuous improvement in our model's fidelity demonstrates the benefit of the proposed mesh update module. 
BundleSDF \cite{bundlesdf} outperforms our method when the number of frames is 16, but its efficiency is notably lower than ours. An analysis is presented in Sec \ref{seclim}.
\begin{table}[t]

\caption{Quantitative comparison with BundleSDF \cite{bundlesdf}.}
\centering
	\resizebox{1.0\columnwidth}{!}{
\begin{tabular}{c|c|c|c|c|c|c|c}
\hline 

\multirow{2}{*}{Object} & \multirow{2}{*}{Method} & \multirow{2}{*}{Metric} & \multicolumn{5}{c}{Number of Frames} \\ \cline{4-8}
 &  & & 1 & 4 & 8 & 12 & 16 \\
\hline
\multirow{2}{*}{$\mathrm{YCB}_{15}$} & \multirow{2}{*}{BundleSDF} &CD $\times 10^{3}$  & 6.19  & 5.21 & 4.27 & 3.60  & \textbf{2.52}  \\ \cline{3-8}
   &  & Time (s) & 110.6  & 114.2 & 123.6 & 135.8  & 142.3  \\ \cline{2-8} 
 Power & \multirow{2}{*}{Ours} &CD $\times 10^{3}$  & \textbf{3.53}  & \textbf{3.44} & \textbf{3.19} & \textbf{3.08}  & 2.97  \\ \cline{3-8}
 Drill &  & Time (s) & 8.1  & 3.1 & 5.8 & 8.7  & 9.6  \\ \cline{2-8}
\hline
\multirow{2}{*}{$\mathrm{YCB}_{5}$} & \multirow{2}{*}{BundleSDF} &CD $\times 10^{3}$  & 5.62  & 4.69 &  3.88 &  3.13 & \textbf{ 2.18}  \\ \cline{3-8}
 &  & Time (s) & 105.5  & 109.3 & 117.2 & 121.4  & 135.5  \\ \cline{2-8} 
 Mustard & \multirow{2}{*}{Ours} &CD $\times 10^{3}$  & \textbf{3.12}  & \textbf{3.01} & \textbf{2.89} & \textbf{2.86}  & 2.77  \\ \cline{3-8}
Bottle &  & Time (s) & 8.1  & 1.8 & 4.6 & 7.3  & 9.3  \\ \cline{2-8}
\hline
 \multirow{2}{*}{$\mathrm{YCB}_{3}$} & \multirow{2}{*}{BundleSDF} &CD $\times 10^{3}$  & 4.48  &  3.52 &  2.75 & 2.27  & \textbf{2.01}  \\ \cline{3-8}
  &  & Time (s) & 102.8  & 109.9 & 123.2 & 133.6  & 136.9  \\ \cline{2-8} 
 Sugar & \multirow{2}{*}{Ours} &CD $\times 10^{3}$  & \textbf{3.10}  & \textbf{3.05} & \textbf{2.68} & \textbf{2.15}  & 2.63  \\ \cline{3-8}
 Box&  & Time (s) & 8.1  & 1.4 & 3.6 & 6.1  & 9.2  \\ \cline{2-8}
 \hline
Real & \multirow{2}{*}{BundleSDF} &CD $\times 10^{3}$  & 7.81  &  6.67 &  4.70 & 3.94  & \textbf{ 2.75}  \\ \cline{3-8}
 World  &  & Time (s) & 103.3  & 210.1 & 296.5 & 334.5  & 453.7  \\ \cline{2-8} 
  \multirow{2}{*}{Mug} & \multirow{2}{*}{Ours} &CD $\times 10^{3}$  & \textbf{3.72}  & \textbf{3.61} & \textbf{3.38} & \textbf{3.22}  & 3.20  \\ \cline{3-8}
 &  & Time (s) & 8.2  & 7.2 & 9.8 & 9.9  & 9.9  \\ \cline{2-8}

\hline 
\end{tabular}
}
\vspace{-0.2in}
\label{sdftab}

\end{table}

\subsection{Computational Time Analysis and Ablation Studies} \label{ab}
\noindent{\textbf{Computational Time Analysis.}} We first report the computational time of the major steps of our method in Table \ref{timetab}, where the results correspond to the tests shown in Table \ref{ycbtab}. As seen, HIPPo can estimate the 6D pose of a novel object at around 15 FPS (object segmentation and 6D pose estimation) after the initialization of HIPPo Dreamer (image-to-multivew-to-mesh), which takes around 8 seconds. The most time-consuming process is the mesh update, and a more detailed analysis of this process is provided in the following ablation studies.
\begin{table}[h]
\caption{Computational Time Analysis of HIPPo.}
\centering
	\resizebox{1.0\columnwidth}{!}{
\begin{tabular}{c|c|c|c|c|c}
\hline 
\multirow{2}{*}{Process} & Object& Image-to- & Multiview-to & 6D Pose  & Mesh\\
 & Segmentation& -Multiview & -Mesh & Estimation & Update\\
\hline
Time (s) & 0.04 & 2.01 & 6.05 & 0.03 & 9.87\\
\hline
\end{tabular}
}
\label{timetab}

\end{table}
\\
\noindent{\textbf{Effect of Mesh Update Frequency.}} We first study the effect of mesh update frequency. In particular, the frequency is controlled by the number of viewpoints on the sphere shown in Fig. \ref{sphere}. The result on YCB-Video \cite{ycbv} is presented in Table \ref{abtab1}. As seen, increasing the mesh update frequency can slightly improve the 6D pose estimation performance. However, considering that mesh update also consumes time, it is not practical to update the mesh with a high frequency in real applications. We choose 36 viewpoints for the balance of accuracy and efficiency. Moreover, one viewpoint on the sphere indicates no mesh update. This is a feasible solution for scenarios where no time is allowed for mesh updates or collecting the object's asset is not necessary. However, Fig. \ref{sdf} demonstrates the necessity of mesh updates for collecting high-fidelity 3D assets.
\begin{table}[h]

\caption{Ablation study on the effect of mesh update frequency.}
\centering
	\resizebox{0.7\columnwidth}{!}{
\begin{tabular}{c|c|c|c|c}
\hline 
Num. of Viewpoint & 1 & 25 & 36 & 64\\
\hline
ADD  & 86.57 & 88.71 & 89.07 & 90.31 \\
\hline
ADD-S  & 94.62 & 96.55 & 97.00 & 97.21\\
\hline 
\end{tabular}
}

\label{abtab1}

\end{table}

\begin{figure}[t]
	\centering
	\includegraphics[width=3.3in]{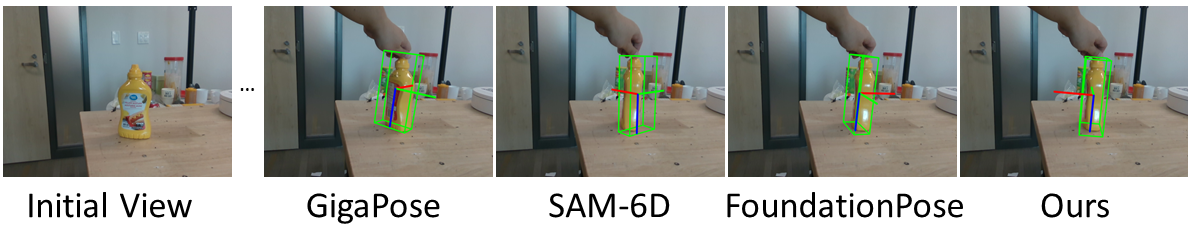}
	\caption{A qualitative comparison with the SOTA methods \cite{gigapose,sam6d,fp}.}
	\label{6ddemo}

\end{figure} 
\begin{figure}[t]
	\centering
	\includegraphics[width=3.3in]{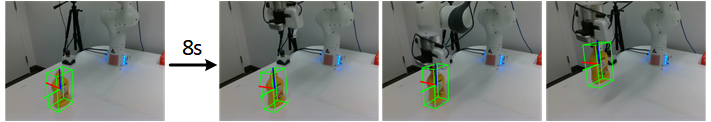}
	\caption{An illustration of HIPPo's immediate robotic application.}
	\label{demo}
\end{figure} 

\noindent{\textbf{Effect of the Number of Points on Mesh Update.}}
We then study the impact of the number of preserved points after downsampling. In Table \ref{abtab2}, we report the effect of the number of preserved measured points after downsampling, as the efficiency of the mesh update is determined by it. We use the mug scenario shown in Fig. \ref{sdf} as an example and focus on the mesh optimization case with 16 frames. As shown, preserving more points, though beneficial for improving reconstruction quality, leads to a longer inference time. Thus, we opt to preserve 30,000 points for the quality-efficiency balance.
\begin{table}[h]

\caption{Ablation study on the effect of the number of preserved points after downsampling for mesh update.}
\centering
	\resizebox{0.8\columnwidth}{!}{
\begin{tabular}{c|c|c|c|c|c}
\hline 
Num. of Points & 5000 & 10,000 & 20,000 & 30,000 & 40,000\\
\hline
Time (s) & 1.73 & 3.52 & 7.26 & 9.87 & 14.34 \\
\hline
CD $\times 10^{3}$ & 3.55 & 3.42 & 3.30 & 3.17 & 3.13 \\
\hline
\end{tabular}
}

\label{abtab2}

\end{table} \\
\noindent{\textbf{Effects of Scale Recovery and the SOR Filter.}} We study the effects of scale recovery, introduced in Sec. \ref{frameworkdesign}, and the SOR filter, which is used for denoising point clouds. The result on YCB-Video \cite{ycbv} is presented in Table \ref{abscale}. As seen, since the correct scale of the reference mesh is a prerequisite for 6D pose estimation, removing both the scale recovery and the SOR filter leads to severe performance degradation. In particular, scale recovery impacts the performance more, as the SOR filter only removes the effect of noise on scale, while scale recovery directly determines the scale. The results demonstrate that both of them are necessary components in HIPPo Dreamer.
\begin{table}[t]

\caption{Ablation study regarding scale recovery and the SOR filter.}
\centering
	\resizebox{1.0\columnwidth}{!}{
\begin{tabular}{c|c|c|c|c}
\hline 
\multirow{2}{*}{Method}  & w/o Scale Recovery & w/o Scale Recovery & w/ Scale Recovery &w/ Scale Recovery\\
 & w/o SOR Filter& w/ SOR Filter &  w/o SOR Filter & w/ SOR Filter\\ \hline
ADD  & 15.77 & 41.20 & 80.87 & 89.07 \\
\hline
ADD-S  & 31.43 & 75.64 & 92.35 & 97.00\\
\hline 
\end{tabular}
}

\label{abscale}

\end{table}

\begin{figure}[t]
	\centering
	\includegraphics[width=3.3in]{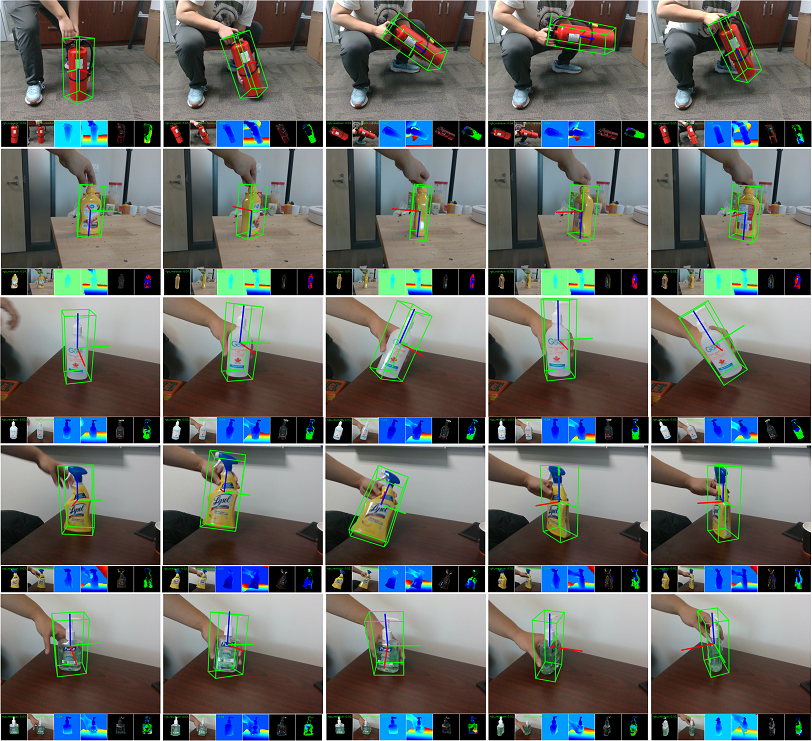}
	\caption{Demonstration of HIPPo's zero-shot pose estimation ability. From top to bottom: Fire Extinguisher, Great Value Mustard, Germs Be Gone Sanitizer, Lysol Wipes, and Purell Sanitizer. At the bottom of each image, six sub-images are displayed: rendered RGB, measured RGB, rendered depth, measured depth, RGB residual, and depth residual.}
   
	\label{any}
\end{figure} 

\subsection{Demonstration of Robotic Application} \label{secdemo}
First, we present a qualitative comparison with the SOTA competitors \cite{fp,sam6d,gigapose} in the following scenario: the robot needs to instantly estimate the 6D pose of a novel object, while movement around the object to observe it is not allowed. Therefore, the only available information is the first RGB-D frame of the object. The SOTA competitors have to utilize the incomplete model of the object built from this single frame. In contrast, HIPPo can instantly construct the complete 3D model by harnessing image-to-3D priors and carry out this task. The qualitative comparison is shown in Fig. \ref{6ddemo}. Then, we present an application demo in Fig. \ref{demo}. The task is to grasp a novel object while also obtaining its 3D oriented bounding box for further planning. The frames are captured by a static calibrated RGB-D camera. If we employ FoundationPose \cite{fp} for this task, we have to reconstruct \cite{bundlesdf} the object first, which could take several minutes. In comparison, by leveraging HIPPo, the robot only needs to wait a few seconds to execute the task. Moreover, thanks to the robust zero-shot prediction ability of the proposed HIPPo Dreamer, as shown in Fig. \ref{any}, our method can estimate the 6D pose of novel real-world objects.

\subsection{Limitations} \label{seclim}
Most image-to-3D methods struggle with severe object occlusion because they only learn in an image-to-image manner \cite{vqadiff}. This affects the 3D mesh generated by HIPPo Dreamer. Due to page limitations, more results are discussed in the supplementary material.
A possible solution to this problem is to apply a Vision-Language model \cite{blip2}, as in \cite{vqadiff}. Efficiency is a major concern when designing HIPPo, so the 3D colored point cloud modality is utilized to develop the mesh optimization algorithm. Consequently, when observations are sufficient and time consumption is not considered, its reconstruction quality is inferior to shape representations \cite{bundlesdf,mvdeepsdf} focusing on rendering quality.
\section{Conclusion}\label{con}
In this work, we propose a new HIPPo framework that harnesses image-to-3D priors for model-free zero-shot 6D pose estimation.
We design a novel instant image-to-mesh strategy, called HIPPo Dreamer, to generate a 3D mesh from the first glance of the object in mere seconds. In addition, we develop a measurement-guided formulation that gradually updates the diffusion prior with more reliable online measurements of geometry and appearance. Compared to existing 6D pose estimation methods, HIPPo does not require a textured 3D model or reference images in advance.
Compared to object SLAM methods, HIPPo always maintains a complete model, supporting immediate robotic applications.
Qualitative and quantitative evaluations on various benchmarks show that the proposed approach outperforms the state-of-the-art 6D pose estimation methods when prior reference images are scarce.

	%

	\bibliographystyle{IEEEtran} 
	\bibliography{reference}

\end{document}